\documentclass[letterpaper, 10 pt, conference]{ieeeconf}  % Comment this line out
\IEEEoverridecommandlockouts                              % This command is only
\usepackage[utf8]{inputenc}
\usepackage[T1]{fontenc}

\usepackage{cite}
\usepackage[pdftex]{graphicx}
\usepackage{amsmath}
\usepackage{amssymb}
\usepackage{array}
\usepackage{wasysym}
\usepackage{bm}
\usepackage{siunitx}
\usepackage{url}
\usepackage{color}
\usepackage{epstopdf}
\usepackage[ruled]{algorithm2e} 
\usepackage{arydshln}
\usepackage{mathtools}

\usepackage{stfloats}
\usepackage{relsize}
\usepackage{cancel}
\usepackage{epstopdf}
\usepackage{environ}
\usepackage{amsthm}
\usepackage{cite}
\usepackage{balance}
\usepackage{hyperref}

\NewEnviron{ruledtable}{%
\par\addvspace{2mm}\hrule height 0.03cm 
\begin{table}[!h]\BODY\end{table}
\hrule height 0.03cm \addvspace{2mm}
}

\DeclareMathOperator*{\argmin}{arg\,min}

\theoremstyle{plain}

\newtheorem{definition}{Definition}

\usepackage{CJKutf8}

\begin{document}
% \title{\LARGE \bf{Unified Framework for Safe-critical Autonomy in Autonomous Inspection using Quadruped Robots Equipped with an Arm}}
% \title{\LARGE \bf{A Unified Framework for Safety-critical Autonomous Inspection \newline
% of Distillation Columns using Quadruped Robots Equipped with Roller Arms}}
\title{\LARGE \bf{Safety-critical Autonomous Inspection 
of Distillation Columns using Quadrupedal Robots Equipped with Roller Arms}}
\author{Jaemin Lee, Jeeseop Kim, and Aaron D. Ames
\thanks{This work is supported by Dow under the project  \#227027AT.}
\thanks{The authors are with the Department of Mechanical and Civil Engineering, California Institute of Technology, Pasadena, CA, USA} 
}

\maketitle

\begin{abstract}
This paper proposes a comprehensive framework designed for the autonomous inspection of complex environments, with a specific focus on multi-tiered settings such as distillation column trays. Leveraging quadruped robots equipped with roller arms, and through the use of onboard perception, we integrate essential motion components including: locomotion, safe and dynamic transitions between trays, and intermediate motions that bridge a variety of motion primitives. Given the slippery and confined nature of column trays, it is critical to ensure safety of the robot during inspection, therefore we employ a safety filter and footstep re-planning based upon control barrier function representations of the environment. Our framework integrates all system components into a state machine encoding the developed safety-critical planning and control elements to guarantee safety-critical autonomy, enabling autonomous and safe navigation and inspection of distillation columns. 
% Key elements of the proposed framework include autonomous transitions between column trays based on measured manway information, safety-critical planning and control through the state machine, and intermediate motions that bridge various motion primitives. 
Experimental validation in an environment, consisting of industrial-grade chemical distillation trays, highlights the effectiveness of our multi-layered architecture.
%, demonstrating successful autonomous and safe inspections in complex real-world scenarios.
\end{abstract}

\section{Introduction}

As robots and their locomotion capabilities continue to advance, they have become integral in various applications such as logistics \cite{echelmeyer2008robotics, lee2023real}, inspection \cite{lattanzi2017review, lim2011developing, lee2023safetyinspection}, and rescue operations \cite{murphy2004human}. Particularly in industrial settings, there is a growing demand for autonomous robotic systems to undertake hazardous missions, replacing human workers in dangerous, complex, and spatially constrained environments. The challenges posed by such environments necessitate a heightened focus on ensuring the safety of robotic deployments, especially in intricate spaces that may contain unsafe regions, exemplified by distillation column trays \cite{molnar2023mechanical,lee2023safetyinspection}.

Autonomous robotic systems are crucial in scenarios where human workers or operators cannot be present, and ensuring the safety of these systems becomes paramount. This paper introduces a comprehensive framework designed for the autonomous inspection of legged robots equipped with roller arms in multi-layered trays, as illustrated in Fig. 1. Building upon the results from our previous work \cite{molnar2023mechanical,lee2023safetyinspection}, we integrate planning and control stacks with a state machine and perception package to enhance the autonomy of the robot. Our proposed architecture enhances the locomotion stack by incorporating considerations for multiple safety conditions within the column tray. Intermediate motions are introduced to bridge the gap between the locomotion stack and transition motion primitives, which is essential for navigating between different column layers. The implemented state machine empowers the robot to autonomously execute inspection missions autonomously, utilizing perceptual information in conjunction with the planning and control stacks. This unified approach ensures seamless planning, control, and perception integration for effective autonomous inspection in complex and confined spaces.

\begin{figure}[t] 
\centering
\includegraphics[width=\linewidth]{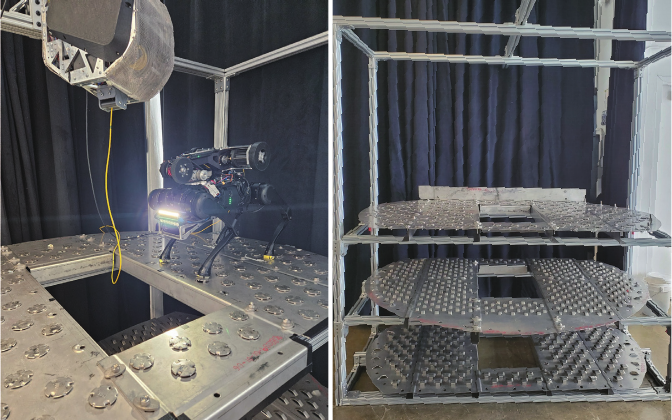}
\vspace{-5mm}
\caption{\textbf{Robotic system and industry-grade column tray}: Our robotic system consists of a quadruped robot, roller arm, perception package, and tether system, as shown in the left figure. }
\vspace{-5mm}
\label{Fig0}
\end{figure}

 \subsection{Related Works}
% safety-critical control
Safety-critical control plays a crucial role in guiding robotic systems, including quadrupeds \cite{lee2023safetyinspection, molnar2021model, grandia2021multi, lee2023hierarchical}, bipedal robots \cite{agrawal2017discrete, csomay2021episodic}, drones \cite{cosner2023generative}, and others. Specifically, control barrier functions (CBFs) \cite{ames2016control} lead to the notion of of safety filters \cite{wabersich2023data}, represented as quadratic programs (QPs), to effectively regulate control inputs and ensure safety. Moreover, CBF-based safety filters have been successfully implemented in robotic systems using reduced-order models like the double integrator \cite{lee2023hierarchical, molnar2021model, kim2023safety, jankovic2023multiagent} or single-rigid body systems. This approach enables the implementation of safety filters through low-dimensional state variables, ensuring efficient computational performance. The application of CBF-based approaches extends beyond generating safe velocity commands for quadruped robots; it also includes the ability to switch gait types in legged robots, enhancing safety during locomotion \cite{lee2023safetyinspection}.

% whole-body control
After obtaining body velocity commands through the aforementioned safety filters, the subsequent step typically involves the utilization of either whole-body control (WBC) \cite{lee2012intermediate, rocchi2015opensot, kim2020dynamic, bellicoso2017dynamic} or inverse dynamics control (IDC) \cite{zapolsky2014quadratic, righetti2011inverse} to compute the actuator commands. To enhance robustness in the face of uncertain disturbances, adaptive approaches \cite{lee2022online, lee2016robust} and task-space decomposition strategies \cite{farshidian2017robust} are commonly employed, bolstering the robot's behaviors. While there is a growing interest in exploring the synergy between whole-body control and learning techniques for simpler tasks like walking \cite{viceconte2022adherent} and push-recovery \cite{ferigo2021emergence} in legged manipulators, sophisticated and consecutive motions often benefit more from a verifiable model-based whole-body control approach, particularly when augmented with a well-designed state machine.

\begin{figure*}[t] 
\centering
\includegraphics[width=0.9\linewidth]{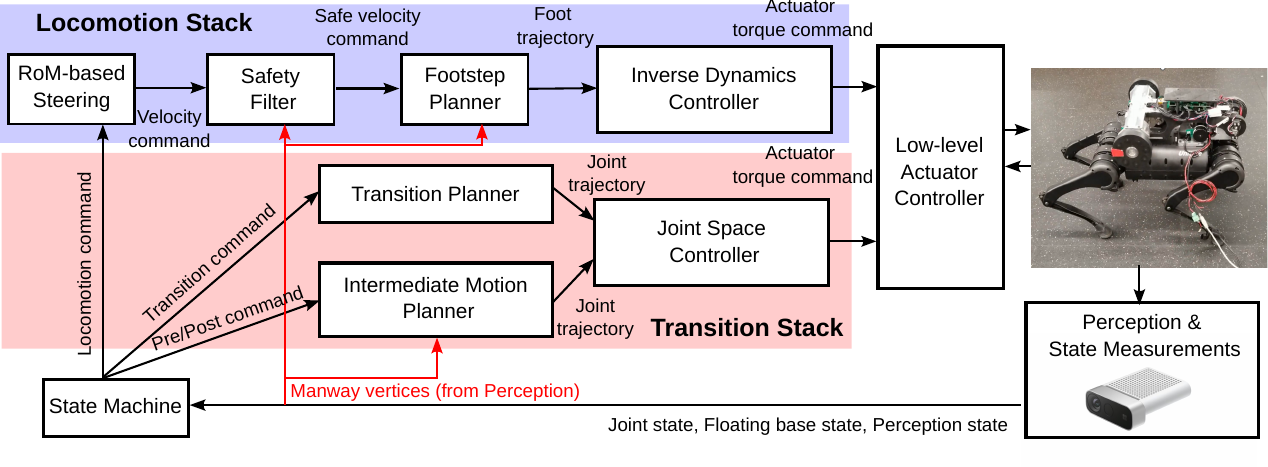}
\vspace{-5mm}
\caption{\textbf{Overall structure of the proposed framework}: The proposed framework consists of a locomotion stack, transition stack, and state machine for autonomy. In addition, the perception part provides the manway vertices to ensure the safety in the planning parts.}
\label{Fig1}
\vspace{-0.5cm}
\end{figure*}

% Perception for locomotion
An additional imperative in safety-critical planning and control involves the integration of a perception stack within a feedback loop. By incorporating the uncertain state data acquired from sensors, advancements have been made in enhancing CBF-based safety filters by utilizing measurement robust CBFs, as demonstrated and validated using Segway in \cite{cosner2021measurement}. In the context of bipedal locomotion, point cloud data is transformed into regions comprising convex polygons, effectively delineating safe and unsafe zones in challenging terrains \cite{bertrand2020detecting, mishra2021gpu}. Quadruped robots similarly leverage perception packages to facilitate more responsive and secure locomotion \cite{havoutis2013onboard, miki2022learning}. As demonstrated in the aforementioned studies, ensuring reliable perception guarantees safety during locomotion in diverse environments, encompassing both the wild outdoors and confined indoor spaces. This becomes particularly critical when controlling a robot within complex and limited spaces, such as our industry-grade column tray.  

% autonomy of the robots
Autonomy stands as a pivotal factor in the successful execution of inspection missions using robotic systems. Without autonomy, manual control by human operators or direct access to monitor the robots' status becomes necessary. Traditionally, the significance of autonomy has been highlighted in scenarios involving multiple robots or agents \cite{goodrich2007managing, schermerhorn2009dynamic}. Even in the context of a single agent, autonomy plays a crucial role in the seamless execution of complex missions guided by a predefined logic, often represented by a control graph of activities, as exemplified in \cite{alami1998architecture}. In this context, practical demonstrations with autonomy are facilitated using behavior trees or finite state machines \cite{iovino2023programming, ghzouli2023behavior}. The key lies in designing the state machine with meticulous consideration of the specific robotic applications and the given missions. 

%%% 

This paper aims to develop a full safety-critical autonomy stack for the inspection of chemical distillation columns.  To this end, we leverage novel hardware consisting of a quadrupedal robot with an attached roller arm \cite{molnar2023mechanical}.  This hardware was designed with a view toward operating in the confined spaces of these column trays and, importantly, transitioning between the trays. Yet utilizing this hardware in an autonomous fashion has yet to be demonstrated.

 \subsection{Contributions}
The primary innovation presented in this paper lies in introducing a comprehensive framework designed for the autonomous inspection of robotic systems illustrated in Fig. \ref{Fig1}. This framework is specifically tailored for deployment in industry-grade trays encompassing multiple columns and consisting of locomotion stack, transition stack, perception part, and state machine, as shown in Fig. \ref{Fig1}. Compared with our prior research efforts in \cite{lee2023safetyinspection}, the specific contributions achieved in this work are outlined as follows:
 \begin{itemize}
     \item A state machine has been specifically crafted to enhance the autonomy of robotic inspection within the distillation column tray.
     \item Intermediate motions are meticulously generated to facilitate smooth transitions between the locomotion and transition stacks.
     \item The incorporation of a perception package enables autonomous navigation, allowing the system to avoid unsafe regions during the execution of safe locomotion intelligently.
     \item Safeguarding the integrity of the columns is prioritized through the simultaneous consideration of two CBFs in the planning process, leveraging a reduced-order model for enhanced efficiency.
     \item The efficacy of the proposed architecture is empirically validated through a series of experiments conducted in an industry-graded environment.
 \end{itemize}

The structure of our paper is as follows. In Section \ref{section2}, we provide a concise overview of rigid-body dynamics, reduced-order models for robotic systems, and the fundamental concept of CBFs. Section \ref{section3} details the proposed framework, encompassing safety-critical planning, footstep replanning, full-body control, intermediate motions, state machines, and perception packages. In Section \ref{section4}, we validate the effectiveness of the proposed framework through autonomous inspection scenarios involving a quadruped robot equipped with a roller arm in an industry-grade column tray.

\section{Preliminaries}
\label{section2}
This section briefly overviews the system models employed in the proposed framework, including both full-body and reduced-order models. In addition, we introduce the fundamental concept of CBFs.

\subsection{System Models}
When considering the full-body dynamics of the legged robots, the equation of motion is expressed as follows:
\begin{equation} \label{rigid}
    \mathbf{D}(\bm{q}) \ddot{\bm{q}} + \mathbf{H}(\bm{q}, \dot{\bm{q}}) = \mathbf{S}^{\top} \bm{\tau} + \mathbf{J}_c(\bm{q})^{\top} \bm{F}_{c}
\end{equation}
where $\bm{q} \in \mathcal{Q}$ and $\bm{\tau} \in \Gamma$ represent the joint variable and the control torque input, respectively, with a configuration space $\mathcal{Q}\subset\mathbb{R}^{n}$ and control input space $\Gamma\subset\mathbb{R}^{n-6}$. $\mathbf{D}(\bm{q})$, and $\mathbf{H}(\bm{q}, \dot{\bm{q}})$ denote the mass/inertia matrix and the sum of Coriolis/centrifugal and gravitational forces, respectively. $\mathbf{S}\in \mathbb{R}^{(n-6)\times n}$ and $\bm{F}_c \in \mathbb{R}^{n_c}$ represent the selection matrix for the actuation joints and the contact forces, where $n_c$ denotes the dimension of the contact forces under a certain supporting phase. $\mathbf{J}_{c}(\bm{q}) \in \mathbb{R}^{n_c \times n}$ is the contact Jacobian associated with the contact force. This full-body dynamics model computes the joint control command given reference trajectories.

Since the above full-body dynamics model is high-dimensional, we have frequently utilized reduced-order models for safety-critical planning, such as the inverted pendulum, lumped mass, double-integrator, etc. Similar to our previous work in \cite{lee2023safetyinspection}, we choose the $2$-dimensional double integrator system with a state $\bm{\varphi} = [\varphi_x, \varphi_y] \in \mathbb{R}^{2}$. We represent the corresponding state-space model as follows:
\begin{equation}
    \dot{\bm{x}}_{\varphi} \coloneqq f_{\varphi}(\bm{x}_{\varphi}) + g_{\varphi}(\bm{x}_{\varphi}) \bm{\nu}
\end{equation}
where $\bm{\nu} \in \mathbb{R}^{2}$ denotes the input of the double-integrator system, which is the velocity command of the robot's base. It is important to figure out the bounds of the velocity commands $\bm{\nu}_{\max}$ and $\bm{\nu}_{\min}$ for realizable safety-critical commands. 

\subsection{Control Barrier Functions}
Let us consider a control affine system described with a state $\bm{x} \in \mathcal{X}$ and control input $\bm{u}\in \mathcal{U}$ as follows:
\begin{equation}
    \dot{\bm{x}} = f(\bm{x}) +  g(\bm{x}) \bm{u}
\end{equation}
where $f$ and $g$ are Lipschitz continuous. With an assumption that a safe set $\mathcal{C} \coloneqq \{\bm{x} \in \mathbb{R}^{n_x}: h(\bm{x}) \geq 0 \}$ is given where $h$ is a continuous and differentiable function, we can define control barrier functions according to \cite{ames2016control}.
\begin{definition}
    Given a safe set $\mathcal{C} = \{\bm{x}\in \mathbb{R}^{n_x}: h(\bm{x}) \geq 0 \}$, $h: \mathbb{R}^{n_x} \to \mathbb{R}$ is a control barrier function (CBF) if there exists a function $\alpha \in \mathcal{K}_{\infty}^{e}$ such that for all $\bm{x} \in \mathcal{C}$: 
    \begin{equation*}
        \sup_{\bm{u} \in \mathcal{U}} \dot{h}(\bm{x}, \bm{u}) \geq - \alpha(h(\bm{x})).
    \end{equation*}
If all states in $\mathcal{C}$ satisfied the above inequality for inputs in $\mathcal{U}$, $\mathcal{C}$ is forward invariant.
\end{definition}
\noindent
In this paper, we employ the defined CBFs to generate safe control velocity of the robot's base based on the perceptual information of the environments by leveraging the double-integrator system.

\section{The Proposed Framework}
\label{section3}

In this section, we explain the meticulous update and specifications applied to the safety-critical planning, footstep replanning, and full-body control modules for the confined space within the column tray. Moreover, we introduce novel components, including intermediate motions facilitating the connection between locomotion and transition stacks, a state machine, and the perception modules, contributing to the comprehensive evolution of our framework.

\begin{figure}[t] 
\centering
\includegraphics[width=\linewidth]{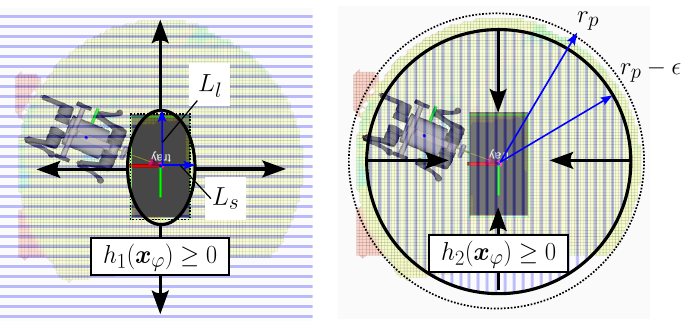}
\vspace{-6mm}
\caption{\textbf{Two CBFs for safety-critical planning in the tray}: $h_1$ and $h_2$ are the CBFs to avoid the manway and the edge of the tray, respectively. }
\label{Fig2}
\vspace{-0.5cm}
\end{figure}

\subsection{Safety-critical Planning}
\label{section3_1}
In the industry-graded column tray, the safe space is limited due to the manway. For this reason, we need to define two safety conditions for ensuring safety in the planning process using the reduced-order model. Let's assume that the manway is rectangular and all vertices are given as $\bm{v}_{1}$, $\bm{v}_{2}$, $\bm{v}_{3}$, and $\bm{v}_{4}$ where the lengths of long and short sides are $L_{l}$ and $L_{s}$, respectively. In addition, the radius of the column plate is $r_p$. First, to avoid the manway, which is an unsafe region, we designed the CBF in an ellipsoidal shape as we proposed in \cite{lee2023safetyinspection}:
\begin{equation}
\begin{split}
      h_1(\bm{x}_{\varphi}) =& \bm{\varphi}^{\top} \mathbf{A} \bm{\varphi} + \mathbf{B}\bm{\varphi} + c.
\end{split}
\end{equation}
The detailed $\mathbf{A}$, $\mathbf{B}$, and $c$ are expressed as follows: 
\begin{equation*}
    \begin{split}
      \mathbf{A} =& \left[\begin{array}{cc}\frac{\cos^2\theta }{a^2} + \frac{\sin^2\theta}{b^2} & \left(\frac{1}{a^2} - \frac{1}{b^2}\right)\cos \theta \sin \theta \\ \left(\frac{1}{a^2} - \frac{1}{b^2}\right)\cos \theta \sin \theta & \frac{\sin^2\theta }{a^2} + \frac{\cos^2\theta}{b^2} \end{array} \right] \\
      \mathbf{B} =& - 2 \overline{\bm{\varphi}}^{\top} \mathbf{A}, \quad  c = \overline{\bm{\varphi}}^{\top} \mathbf{A} \overline{\bm{\varphi}} -1
    \end{split}
\end{equation*}
where $a = L_l + \varepsilon_{l}$, $b = L_s + \varepsilon_{s}$, and $\theta$ are the lengths of the major and minor radii and the orientation of the ellipse. In addition, $\overline{\bm{\varphi}}$ denotes the center of the manway. Second, the robot's base must be inside the column plates. For simplicity, we can define the CBF for this condition as follows:
\begin{equation}
    h_2(\bm{x}_{\varphi}) = -(\bm{\varphi} - \overline{\bm{\varphi}})^{\top}(\bm{\varphi} - \overline{\bm{\varphi}}) + (r_p - \epsilon)^{2}
\end{equation}
where $\epsilon$ denotes the offset for considering the body frame of the robot's base.

Based on the above two CBFs, a QP problem is formulated to ensure the safety of the robotic systems as follows:
\begin{equation}\label{QP1}
    \begin{split}
        \argmin_{\bm{\nu} \in \mathcal{V} } & \quad \| \bm{k}^{d}(\bm{x}_{\varphi}, \bm{\xi}) - \bm{\nu} \|^{2}, \\
        \textrm{s.t.} &\quad \dot{h}_{1}(\bm{x}_{\varphi}, \bm{\nu}) \geq - \alpha_{1}(h_{1}(\bm{x}_{\varphi})), \\
        &\quad \dot{h}_{2}(\bm{x}_{\varphi}, \bm{\nu}) \geq - \alpha_{2}(h_{2}(\bm{x}_{\varphi}))
    \end{split}
\end{equation}
where $\bm{k}^{d}: \mathcal{X}_{\varphi} \times \mathbb{R}^{4} \to \mathcal{V}$ is the controller to track the desired position $\bm{\xi}$. The reference controller can be designed as a simple P controller $\bm{k}^{d}(\bm{x}_{\varphi})= \mathbf{K}_{\varphi}(\bm{\xi} - \bm{x}_{\varphi})$ where $\mathbf{K}_{\varphi} = \textrm{diag}(k_{\varphi_x}, k_{\varphi_y})$ with the proportional gains in the $x$ and $y$ direction, $k_{\varphi_x}$ and $k_{\varphi_y}$. $\alpha_1$ and $\alpha_2$ denote the $\mathcal{K}_{\infty}^{e}$ functions associated with $h_1$ and $h_2$, respectively. In this work, we assume that the goal positions or waypoints of the robot base are reachable. Otherwise, we need to modify $L_{l}$ and $L_{s}$ to adjust the CBF $h_1$ or relax one of the conditions as \cite{lee2023hierarchical} to make the QP \eqref{QP1} feasible.   

\subsection{Footstep Replanning}
\label{section3_2}
Since the column tray has really limited space, we implement a quasi-static walking gait, which involves only one leg move, and the rest of the feet maintain solid contact during the swing phase. The walking gait is composed of four phases, $w_{\textrm{FL}}$, $w_{\textrm{BR}}$, $w_{\textrm{FR}}$, and $w_{\textrm{BL}}$ whose the swing foot is front left, back right, front right, and back left. We repeat iterating the phases in a sequence:
\begin{equation}
 w_{\textrm{FL}} \to  w_{\textrm{BR}}\to w_{\textrm{FR}} \to w_{\textrm{BL}} \to w_{\textrm{FL}} \to \cdots   
\end{equation}
while executing the quasi-static walking. Based on the current position and orientation of the robot's base, it is possible to compute the goal position of the swing foot $\bm{p}_{s}^{g}$ with the Raibert heuristics. If the goal position of swing foot $\bm{p}_{s}^{g}$ is near the manway or inside the manway, the position is modified according to \cite{lee2023safetyinspection}. 

In addition, we modified the goal position of the swing foot when it is near to the edge of the column tray or outside the tray. The modified goal position $\bm{p}_{s}^{n}$ is obtained by scaling the position vector with respect to the center of the tray:
\begin{equation}
    \bm{p}_{s}^{n} = \overline{\bm{x}}_{\varphi} + \frac{r_{p}- \epsilon}{\|\bm{p}_{s}^{g} - \overline{\bm{x}}_{\varphi}\|} (\bm{p}_{s}^{g} - \overline{\bm{x}}_{\varphi}).
\end{equation}
It is important that the support polygon with the replanned foot location is conservative to maintain the CoM inside it while executing the next swing phase. In addition, we tested the modified foot location to be kinematically feasible within ROM. After that, the swing foot trajectories are designed by using cubic interpolation with the swing time interval.  

\subsection{Full-body Control}
\label{section3_3}
In this paper, we leverage two types of full-body control employed for locomotion control and joint-space control during the transition and intermediate motions. First, the locomotion controller is formulated in a QP problem. After computing joint configuration and velocity, $\bm{q}^{d}$ and $\dot{\bm{q}}^{d}$, based on the swing foot reference trajectories by solving an inverse kinematics problem, we can formulate the inverse dynamic control QP as follows:
\begin{equation} \label{QP_FBC}
    \begin{split}
        \min_{\ddot{\bm{q}}, \bm{\tau}, \bm{F}_{c}} &\quad \|\ddot{\bm{q}}^{d} - \ddot{\bm{q}}\|_{\mathbf{W}_{\ddot{q}}}^{2}  + \|\bm{\tau}\|_{\mathbf{W}_{\tau}}^{2} + \| \bm{F}_{c}\|_{\mathbf{W}_{c}}^{2} \\
        \textrm{s.t.} &\quad     \mathbf{D}(\bm{q}) \ddot{\bm{q}} + \mathbf{H}(\bm{q}, \dot{\bm{q}}) = \mathbf{S}^{\top} \bm{\tau} + \mathbf{J}_c(\bm{q})^{\top} \bm{F}_{c}, \\
        &\quad \mathbf{J}_{c}(\bm{q})\ddot{\bm{q}} + \dot{\mathbf{J}}_{c}(\bm{q}, \dot{\bm{q}})\dot{\bm{q}} = \bm{0}, \\
        &\quad \bm{F}_{c} \in \mathcal{FC},\quad \bm{\tau} \in \Gamma,\\
        &\quad \ddot{\bm{q}}^{d} = \mathbf{K}_p(\bm{q}^{d} - \bm{q}) + \mathbf{K}_{d}(\dot{\bm{q}}^{d} - \dot{\bm{q}})
    \end{split}
\end{equation}
where $\mathbf{K}_{p}$ and $\mathbf{K}_{d}$ are proportional and derivative gains to design the nominal joint acceleration $\ddot{\bm{q}}^{d}$. In addition, $\mathbf{W}_{\ddot{q}}$, $\mathbf{W}_{\tau}$, and $\mathbf{W}_{c}$ denote weighting matrices assigned to the terms associated with joint acceleration error, control torque input, and contact force, respectively. $\mathcal{FC}$ is the contact wrench cone for the given contacts. 

\begin{figure*}[t] 
\centering
\includegraphics[width=\linewidth]{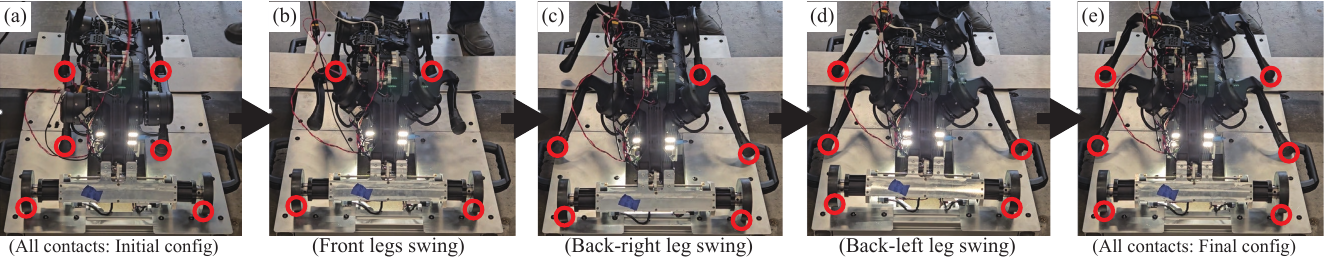}
\vspace{-6mm}
\caption{\textbf{Experimental result generating Pre-motion for downwawrd transition}: By strategically planning intermediate motions prior to initiating the downward transition, the sequence of contact points ensures smooth joint configuration adjustments for the transition. CoM is maintained within the support polygon throughout these intermediate motions.}
\label{Fig3_1}
\vspace{-5mm}
\end{figure*}

\begin{figure}[t] 
\centering
\includegraphics[width=\linewidth]{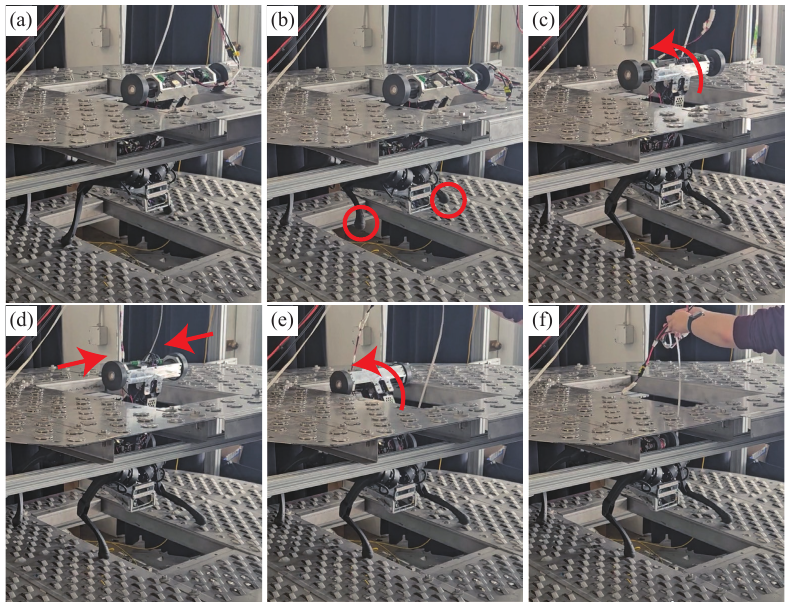}
\vspace{-4mm}
\caption{\textbf{Experimental result generating Post-motion for downward transition}: (a) final configuration post downward transition, (b) front legs realigned to adjust the support polygon, (c) arm slightly folded to detach from the upper layer, (d) arm length adjustment, (e) arm fully folded in preparation for subsequent locomotion, (f) configuration for next locomotion.}
\label{Fig3_2}
\vspace{-0.5cm}
\end{figure}

By using the optimal solution to the above problem $\ddot{\bm{q}}^{\star}$, we can compute the reference joint position and velocity, $\bm{q}^{n}$ and $\dot{\bm{q}}^{n}$, with Euler integration as presented in \cite{lee2023safetyinspection}. Then, the following joint impedance controller is utilized to compute the command torque.
\begin{equation} \label{impedance}
    \bm{\tau}_{\textrm{cmd}} = \bm{\tau}^{\star} + \mathbf{K}_{p}^{\textrm{imp}} ( \bm{q}^{n} - \bm{q}) + \mathbf{K}_d^{\textrm{imp}}(\dot{\bm{q}}^{n} - \dot{\bm{q}})
\end{equation}
where $\mathbf{K}_{p}^{\textrm{imp}}$ and $\mathbf{K}_{d}^{\textrm{imp}}$ represent the proportional and derivative gains for the joint impedance controller. $\bm{\tau}^{\star}$ denotes the optimal solution to the QP problem in \eqref{QP_FBC}. For the transition stack, we make the full-body controller much simpler by replacing $\bm{q}^{n}$ and $\dot{\bm{q}}^{n}$ to $\bm{q}^{d}$ and $\dot{\bm{q}}^{d}$ \eqref{impedance}. In addition, the gravity compensation torque $\bm{\tau}_g$ is plugged into the feedforward torque command instead of $\bm{\tau}^{\star}$. Since the transition and intermediate motions, such as locomotion, are not dynamic, this simplified version of the full-body controller is fairly effective in tracking the reference trajectories. 

\subsection{Intermediate Motions}
\label{section3_4}

Since the transition trajectories are designed offline with respect to a specific joint configuration, it is necessary to change the configuration smoothly by changing the contact sequence. Let's suppose that $N =\{ 1, \cdots, \ell\}$ is the discredited horizon and $\bm{q}_{t}$ is the ready configuration for the transition. Then, we can formulate the following mixed-integer quadratic program (MIQP):
\begin{equation} \label{MIQP}
    \begin{split}
        \min_{\mathbf{C}_{[1:\ell]}, \bm{q}_{[1:\ell]}}  &\quad  (\bm{q}_{t} - \bm{q}^{(\ell)})^{\top} \mathbf{Q}_{q} (\bm{q}_{t} - \bm{q}^{(\ell)}) \\
        \textrm{s.t.}  &\quad \sum_{i=1}^{5} \bm{c}_{i}^{(1)} = 0 , \quad \sum_{i=1}^{5} \bm{c}_{i}^{(\ell)} = 0, \\
        &\quad \sum_{i=1}^{5} \bm{c}_{i}^{(k)}= 2,  \quad k\in\{2, \cdots, \ell-1\},  \\
        &\quad \bm{c}_{1}^{(j)},\: \bm{c}_{4}^{(j)}, \: \bm{c}_{5}^{(j)} \in  \{0,\: 2\}, \quad \bm{c}_{2}^{(j)}, \: \bm{c}_{3}^{(j)} \in  \{ 0, \: 1\}, \\
        &\quad \bm{c}_{i}^{(j)} \mathbf{S}_{i} (\bm{q}_{t} - \bm{q}^{(j)}) = \bm{0}, \quad i \in \{1, \cdots, 5\} 
    \end{split}
\end{equation}
where $\bm{q}^{(j)}$ denotes the $j$-th configuration. The decision parameters $\bm{c}^{(j)} = [\bm{c}_{1}^{(j)},\: \bm{c}_{2}^{(j)},\: \bm{c}_{3}^{(j)},\: \bm{c}_{4}^{(j)}, \:\bm{c}_{5}^{(j)}]$ is defined for the $j$-th step where $\bm{c}_{1}$, $\bm{c}_{2}$, $\bm{c}_{3}$, $\bm{c}_{4}$, and $\bm{c}_{5}$ indicate the integers for contacts of wheel, front-left, front-right, back-left, and back-right legs, respectively.  In addition, $\mathbf{C}_{[1:\ell]} = \{ \bm{c}^{(1)}, \cdots, \bm{c}^{(\ell)}\}$, $\bm{q}_{[1:\ell]} = \{\bm{q}^{(1)}, \cdots, \bm{q}^{(\ell)} \} $ and $\mathbf{S}_i$ is the selection matrix for the leg associated with $\bm{c}_{i}$. The meaningful solution is $\mathbf{C}_{[1:\ell]}$, which is a contact sequence for the intermediate motions. 

Once the optimal contact sequence of the intermediate motions is obtained, we generate continuous trajectories to reach the initial continuation of transition or locomotion by avoiding unexpected collisions. The intermediate motion before the transition requires control of the roller arm to make stable contact with the tray before making sequential motions determined by \eqref{MIQP}. Fig. \ref{Fig3_1} shows the experimental validation of intermediate motions before making the downward transition behavior. Based on the optimal contact sequence, the CoM remains inside of support polygon while changing configuration to ready the downward transition. 

After making the downward transition behavior, we need to consider additional constraints to be ready for next locomotion tasks. The arm must be folded while controlling the CoM inside of the support polygon. As shown in Fig. \ref{Fig3_2}, if needed, we change the foot locations to enlarge the support polygon. These processes are helpful to make smooth swapping between the locomotion and transition stacks. 

\begin{figure}[t] 
\centering
\includegraphics[width=\linewidth]{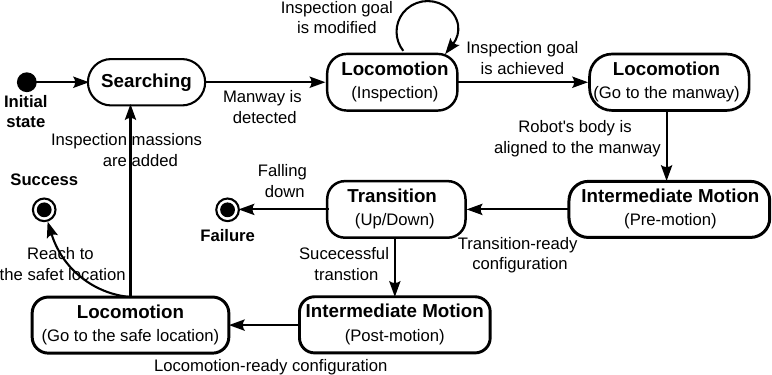}
\vspace{-4mm}
\caption{\textbf{State machine for autonomy while performing inspection missions}: From the initial state to the completion of inspection. In the event of failed transitions, the state machine halts the robot's actions.}
\label{Fig4}
\vspace{-6mm}
\end{figure}

\subsection{State Machine}
\label{section3_5}
To enhance the autonomy of robots within a column tray, we introduce a state machine based on both state and perception measurements. Seven distinct tasks are defined to efficiently execute inspection missions as follows:
\begin{itemize}
    \item \textbf{Searching}: The robot adjusts its body orientation and height at the initial position to obtain the manway vertices.
    \item \textbf{Locomotion} (Inspection): Safely navigating the robot through a single column, ensuring the completion of the inspection mission while avoiding unsafe regions.
    \item \textbf{Locomotion} (Go to the manway): Directing the robot to the manway in preparation for an upward or downward transition.
    \item \textbf{Intermediate Motion} (Pre-motion): Modifying the robot's configuration to prepare for a transition within a specified contact sequence.
    \item \textbf{Transition} (Up/Down): Executing a smooth and safe transition between columns.
    \item \textbf{Intermediate Motion} (Post-motion): Adjusting the robot's configuration to be ready for locomotion again after completing a transition.
    \item \textbf{Locomotion} (Go to the safe location): Navigating the robot to a pre-defined safe location.
\end{itemize}

These tasks form the basis for a defined state machine, as depicted in Fig. \ref{Fig4}. Starting from the initial state, the robot autonomously inspects a single-column tray, capable of repeating the inspection task when goals are added or modified. Following completion of the inspection, the robot positions itself on the manway for upward or downward transitions. Trajectories for the transitions between the column trays are generated by the optimization approach proposed in \cite{molnar2023mechanical} with additional CoM constraints for enhancing the balance during the transitions \cite{lee2023safetyinspection}. It is terminated if a stable transition is not achieved during the process. Upon successful transition, the robot readjusts its configuration for subsequent locomotion. The process concludes if the robot reaches the pre-defined safe location or repeats based on pre-defined conditions. This state machine significantly enhances the autonomy of robotic inspection within multi-column trays, minimizing the need for human intervention. 

\subsection{Perception in a Loop}
\label{section3_6}
Utilizing a Kinect package, we implement a perception stack to discern the manway within the column tray. The primary function of this perception stack is to furnish crucial data regarding the vertices of the manway, denoted as $\bm{v}_1$, $\bm{v}_2$, $\bm{v}_3$, and $\bm{v}_4$. To optimize efficiency and mitigate computation delays in the planning and control loops, we meticulously specify the locations where perceptual data is acquired.

To enhance the precision of our measurements, we conduct 100 data collections and leverage their average. Assuming the relative positions of the manway vertices, $\bm{v}_1^{\{P\}}$, $\bm{v}_2^{\{P\}}$, $\bm{v}_3^{\{P\}}$, and $\bm{v}_4^{\{P\}}$, are determined with respect to the perception frame, we convert these positions into the global coordinate system:
\begin{equation}
\bm{v}_{k}^{\{G\}} = \mathbf{R}_{\{P|G\}} \bm{v}_{k}^{\{P\}} + \bm{\varphi}_{p}^{\{G\}}, \quad \forall k \in \{1,2,3,4\}
\end{equation}
where $\bm{v}_{k}^{\{G\}}$, $\mathbf{R}_{\{P|G\}}$, and $\bm{\varphi}_{p}^{\{G\}}$ represent the position of the $k$-th vertex in the global coordinate system, the rotation matrix from the global to perception coordinates, and the position of the Kinect package with respect to the global frame, respectively.

Based on the obtained vertices $\bm{v}_{k}^{\{G\}}$, we compute the base's velocity command and determine the landing location of the swing foot, as detailed in Sections \ref{section3_1} and \ref{section3_2}. The center of the polygon formed by these vertices is considered as the center of the column tray. Additionally, we make the assumption that given the radius of the column layer and the dimensions of the manway, it is feasible to ascertain the accuracy of the perceptual information.

\section{Experiments}
\label{section4}

\begin{figure*}[t] 
\centering
\includegraphics[width=\linewidth]{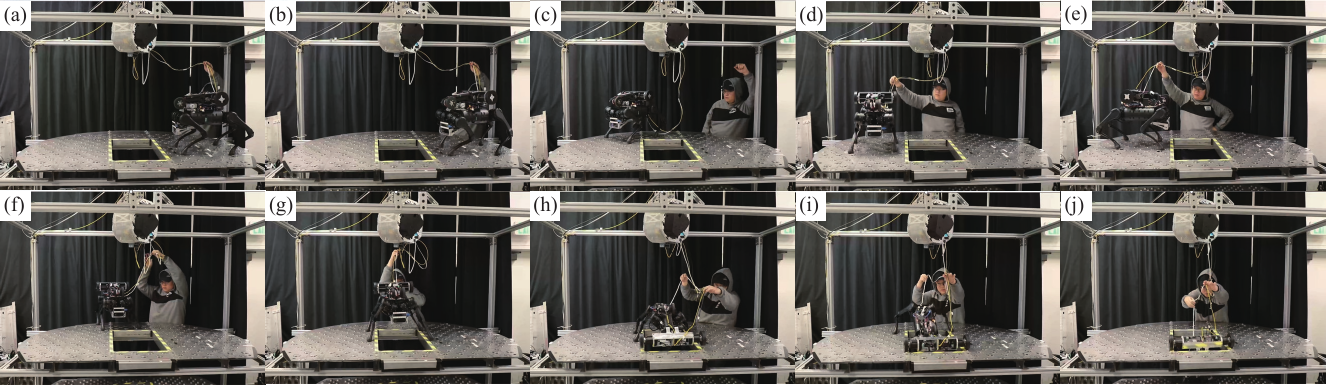}
\vspace{-5mm}
\caption{\textbf{Snapshots of Autonomous Inspection}: (a) initial position establishment, (b) manway search, (c) locomotion execution (inspection) towards designated location, (d) inspection task completion, (e) resuming manway search, (f) moving towards waypoint, (g) approach to transition-ready position on the manway, (h) intermediate motion for downward transition preparation, (i) downward transition execution, (j) completion of downward transition.}
\label{Fig4_1}
\vspace{-0.6cm}
\end{figure*}

In this section, we show the results of our experiments demonstrating the autonomous inspection capabilities of the Unitree A1 quadruped robot equipped with a roller arm within an industry-grade column tray. The roller arm, boasting $5$ DOFs, incorporates $4$ actuators and a bidirectional lead screw for arm extension. Experimental setups employed Intel NUC and Nvidia Jetson AGX Xavier devices, connected to the robot via network switches, to execute and demonstrate the experiments.

Concurrently, trajectory computations for transition and intermediate motions were carried out using FROST \cite{hereid2017frost} and MATLAB on a laptop with a $3.4$ GHz Intel Core i$7$ processor. The optimization solver OSQP \cite{osqp} was employed for solving the QP problem of safety-critical planning and full-body control, executed on the Intel NUC. This integrated approach ensured efficient planning and control mechanisms throughout the inspection process.

The column tray is structured with three layers, each separated by a $22$-inch gap, with a tray radius of $35$ inches. The manway, a crucial component, features dimensions of $27.5$ inches in length and $15$ inches in width. Notably, each layer of the tray is distinguished by unique stud patterns.

\subsection{Autonomous Inspection}

The overall inspection scenario is depicted in Fig. \ref{Fig4_1}. We describe the detailed inspection process corresponding to the snapshot. Initially, the robot is positioned arbitrarily on the top layer of the column tray for autonomous inspection, facing the manway (Fig. \ref{Fig4_1}(a)). To conduct the search motion, we control the robot's body yaw angle within a range of $\pm 0.3$ radians for a duration of $2$ seconds. If perceptual data (vertices of the manway) is not acquired, the robot repeats the search motion until the vertices data is received from the perception stack (Fig. \ref{Fig4_1}(b)). Subsequently, the robot engages in \textbf{Locomotion} (inspection) to reach the predetermined position on the opposite side (Fig. \ref{Fig4_1}(c)). Throughout the inspection task, both CBF constraints $h_1$ and $h_2$ are considered in our locomotion control alongside the footstep re-planning.

Once the inspection task is completed (Fig. \ref{Fig4_1}(d)), the robot is directed towards the manway and repeats the search motion to acquire perception data for the manway (Fig. \ref{Fig4_1}(e)). The \textbf{Locomotion} (Go to the manway) phase comprises two stages: movement towards the waypoint with orientation alignment (Fig. \ref{Fig4_1}(f)) and approach to the transition-ready position on the manway (Fig. \ref{Fig4_1}(g)). While navigating towards the waypoint, the CBF constraints persist to prevent unsafe motions within the manway by ensuring safe foot placement. However, as the robot approaches the transition-ready location, the CBF constraints are lifted, activating only the footstep re-planning component.

Following this, \textbf{Intermediate motion} (Pre-motion) is executed to align the joint configuration with the initial value of the transition trajectory (Fig. \ref{Fig4_1}(h)), which is same as the process illustrated in Fig. \ref{Fig3_1}. The robot then moves to the lower column tray through \textbf{Transition} (Down), as depicted in Fig. \ref{Fig4_1}(i) and (j). In the event of a successful transition, the robot proceeds to execute \textbf{Intermediate motion} (Post-motion), described in Fig. \ref{Fig3_2}. However, if the transition fails, the robot halts, and an additional tether system is employed to pull the robot back to the upper column layer.

\begin{figure}[t] 
\centering
\includegraphics[width=\linewidth]{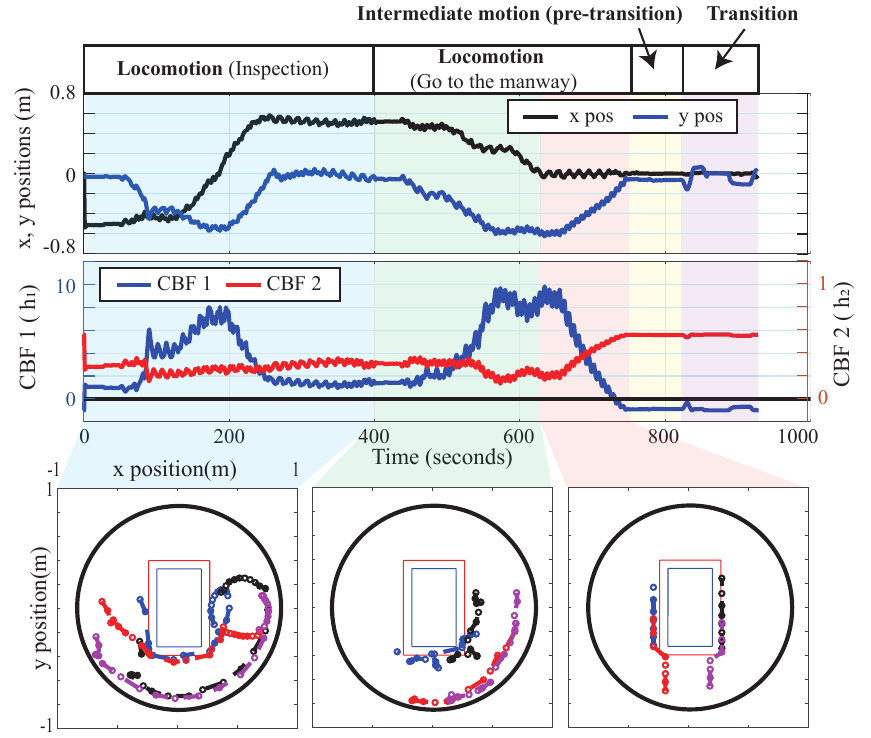}
\vspace{-6mm}
\caption{\textbf{Experimental results}: In the upper portion, color-coded zones delineate distinct stages of locomotion: Blue for inspection, green for approaching a waypoint, red for transition-ready location, yellow for pre-motion before downward transition, and purple for the downward transition. In the lower section, circular markers denote the positions of the front-left (blue), front-right (black), back-left (red), and back-right (purple) feet.}
\label{Fig5}
\vspace{-5mm}
\end{figure}

\subsection{Discussion}
Fig. \ref{Fig5} presents the experimental results of autonomous inspection conducted within the industry-grade column tray. In the upper segment of Fig. \ref{Fig5}, the robot's base position is precisely controlled through full-body control within the confined space without any jerky or unstable motions. In addition, the CBF values for $h_1$ and $h_2$ are always greater than $0$ within the blue and green zones, ensuring the robot's safety. Since we remove the safety filter within the red, yellow, and purple zones, negative CBF value ($h_1$) does not mean any safety violation.

Furthermore, the lower portion of Fig. \ref{Fig5} shows the footstep planning during the inspection. To make footstep re-planning more conservative, we expand the unsafe region based on the geometry of the manway (blue box) and include a $5$ cm buffer margin (red box). All foot placements during autonomous inspection are located within the safe region delineated by the boundary of the column layer (black circle) and the red box, excluding during transition behavior. These results distinctly show that the proposed framework enables safety-critical autonomy of our robotic system within an industry-grade column tray.

However, in the hardware experiments, challenges arose in locomotion control due to the uncertain friction coefficient of the column tray. Initially, we adopted a baseline value of $\mu = 0.7$, which had proven effective for locomotion on flat ground. However, the column tray is much more slippery than the flat ground so we need to tune parameter, resulting in a revised value of $\mu = 0.4$ after iterative hardware experiments. Moreover, the limitation of space rendered the trot gait impractical, leading us to employ the quasi-static gait. Nonetheless, in scenarios where the column tray provides more wider space, employing the trot gait could be feasible instead of the quasi-static gait. To enhance the alignment of transition-ready positions and generate safer motions, it is crucial to achieve more accurate and frequent updates of the perception stack.

\section{Conclusion}
This paper proposes a comprehensive framework for the autonomous inspection of a quadruped robot equipped with a roller arm within an industry-grade column tray featuring multiple layers. Building upon our prior work, we achieve full autonomy in executing inspection missions by employing a state machine to determine a sequence of the robot's behaviors. Additionally, contact sequences for intermediate motions, connecting locomotion and transition stacks, are derived through the solution of a MIQP problem. The manway and tray dimensions are accurately measured using a perception package, and this perceptual information is leveraged to tailor the safety-critical planning and footstep replanning components for operation in confined spaces.

Looking ahead, our future plans involve integrating generative AI with the proposed safety-critical framework to enhance autonomous inspection capabilities across diverse environments. Rather than manually specifying state machines or behavior trees, our goal is to develop a wrapper that bridges the outputs of generative AI with the framework, adapting to dynamic environments that may involve moving objects or unexpected events.

\section*{Acknowledgement}
The authors express their gratitude to collaborators at NASA JPL, particularly for their contributions to the perception component, and the development of the roller arm used throughout the experiments (as documented in \cite{molnar2023mechanical}). 

\bibliographystyle{IEEEtran}
\balance
\bibliography{l_css}

\begin{thebibliography}{10}
\providecommand{\url}[1]{#1}
\csname url@samestyle\endcsname
\providecommand{\newblock}{\relax}
\providecommand{\bibinfo}[2]{#2}
\providecommand{\BIBentrySTDinterwordspacing}{\spaceskip=0pt\relax}
\providecommand{\BIBentryALTinterwordstretchfactor}{4}
\providecommand{\BIBentryALTinterwordspacing}{\spaceskip=\fontdimen2\font plus
\BIBentryALTinterwordstretchfactor\fontdimen3\font minus \fontdimen4\font\relax}
\providecommand{\BIBforeignlanguage}[2]{{%
\expandafter\ifx\csname l@#1\endcsname\relax
\typeout{** WARNING: IEEEtran.bst: No hyphenation pattern has been}%
\typeout{** loaded for the language `#1'. Using the pattern for}%
\typeout{** the default language instead.}%
\else
\language=\csname l@#1\endcsname
\fi
#2}}
\providecommand{\BIBdecl}{\relax}
\BIBdecl

\bibitem{echelmeyer2008robotics}
W.~Echelmeyer, A.~Kirchheim, and E.~Wellbrock, ``Robotics-logistics: Challenges for automation of logistic processes,'' in \emph{IEEE International Conference on Automation and Logistics}, 2008, pp. 2099--2103.

\bibitem{lee2023real}
J.~Lee, M.~Seo, A.~Bylard, R.~Sun, and L.~Sentis, ``Real-time model predictive control for industrial manipulators with singularity-tolerant hierarchical task control,'' in \emph{Proceedings of the IEEE International Conference on Robotics and Automation}, 2023, pp. 12\,282--12\,288.

\bibitem{lattanzi2017review}
D.~Lattanzi and G.~Miller, ``Review of robotic infrastructure inspection systems,'' \emph{Journal of Infrastructure Systems}, vol.~23, no.~3, p. 04017004, 2017.

\bibitem{lim2011developing}
R.~S. Lim, H.~M. La, Z.~Shan, and W.~Sheng, ``Developing a crack inspection robot for bridge maintenance,'' in \emph{Proceedings of the IEEE International Conference on Robotics and Automation}, 2011, pp. 6288--6293.

\bibitem{lee2023safetyinspection}
J.~Lee, J.~Kim, W.~Ubellacker, T.~G. Molnar, and A.~D. Ames, ``Safety-critical control of quadrupedal robots with rolling arms for autonomous inspection of complex environments,'' \emph{arXiv preprint arXiv:2312.07778}, 2023.

\bibitem{murphy2004human}
R.~R. Murphy, ``Human-robot interaction in rescue robotics,'' \emph{IEEE Transactions on Systems, Man, and Cybernetics, Part C (Applications and Reviews)}, vol.~34, no.~2, pp. 138--153, 2004.

\bibitem{molnar2023mechanical}
T.~G. Molnar, K.~Tighe, W.~Ubellacker, A.~Kalantari, and A.~D. Ames, ``Mechanical design, planning, and control for legged robots in distillation columns,'' \emph{Journal of Computational and Nonlinear Dynamics}, vol.~18, no.~6, p. 061001, 2023.

\bibitem{molnar2021model}
T.~G. Molnar, R.~K. Cosner, A.~W. Singletary, W.~Ubellacker, and A.~D. Ames, ``Model-free safety-critical control for robotic systems,'' \emph{IEEE robotics and automation letters}, vol.~7, no.~2, pp. 944--951, 2021.

\bibitem{grandia2021multi}
R.~Grandia, A.~J. Taylor, A.~D. Ames, and M.~Hutter, ``Multi-layered safety for legged robots via control barrier functions and model predictive control,'' in \emph{2021 IEEE International Conference on Robotics and Automation (ICRA)}.\hskip 1em plus 0.5em minus 0.4em\relax IEEE, 2021, pp. 8352--8358.

\bibitem{lee2023hierarchical}
J.~Lee, J.~Kim, and A.~D. Ames, ``Hierarchical relaxation of safety-critical controllers: Mitigating contradictory safety conditions with application to quadruped robots,'' \emph{arXiv preprint arXiv:2305.03929}, 2023.

\bibitem{agrawal2017discrete}
A.~Agrawal and K.~Sreenath, ``Discrete control barrier functions for safety-critical control of discrete systems with application to bipedal robot navigation.'' in \emph{Robotics: Science and Systems}, vol.~13.\hskip 1em plus 0.5em minus 0.4em\relax Cambridge, MA, USA, 2017, pp. 1--10.

\bibitem{csomay2021episodic}
N.~Csomay-Shanklin, R.~K. Cosner, M.~Dai, A.~J. Taylor, and A.~D. Ames, ``Episodic learning for safe bipedal locomotion with control barrier functions and projection-to-state safety,'' in \emph{Learning for Dynamics and Control}.\hskip 1em plus 0.5em minus 0.4em\relax PMLR, 2021, pp. 1041--1053.

\bibitem{cosner2023generative}
R.~K. Cosner, I.~Sadalski, J.~K. Woo, P.~Culbertson, and A.~D. Ames, ``Generative modeling of residuals for real-time risk-sensitive safety with discrete-time control barrier functions,'' \emph{arXiv preprint arXiv:2311.05802}, 2023.

\bibitem{ames2016control}
A.~D. Ames, X.~Xu, J.~W. Grizzle, and P.~Tabuada, ``Control barrier function based quadratic programs for safety critical systems,'' \emph{IEEE Transactions on Automatic Control}, vol.~62, no.~8, pp. 3861--3876, 2016.

\bibitem{wabersich2023data}
K.~P. Wabersich, A.~J. Taylor, J.~J. Choi, K.~Sreenath, C.~J. Tomlin, A.~D. Ames, and M.~N. Zeilinger, ``Data-driven safety filters: Hamilton-jacobi reachability, control barrier functions, and predictive methods for uncertain systems,'' \emph{IEEE Control Systems Magazine}, vol.~43, no.~5, pp. 137--177, 2023.

\bibitem{kim2023safety}
J.~Kim, J.~Lee, and A.~D. Ames, ``Safety-critical coordination for cooperative legged locomotion via control barrier functions,'' \emph{arXiv preprint arXiv:2303.13630}, 2023.

\bibitem{jankovic2023multiagent}
M.~Jankovic, M.~Santillo, and Y.~Wang, ``Multiagent systems with cbf-based controllers: Collision avoidance and liveness from instability,'' \emph{IEEE Transactions on Control Systems Technology}, 2023.

\bibitem{lee2012intermediate}
J.~Lee, N.~Mansard, and J.~Park, ``Intermediate desired value approach for task transition of robots in kinematic control,'' \emph{IEEE Transactions on Robotics}, vol.~28, no.~6, pp. 1260--1277, 2012.

\bibitem{rocchi2015opensot}
A.~Rocchi, E.~M. Hoffman, D.~G. Caldwell, and N.~G. Tsagarakis, ``Opensot: a whole-body control library for the compliant humanoid robot coman,'' in \emph{2015 IEEE International Conference on Robotics and Automation (ICRA)}.\hskip 1em plus 0.5em minus 0.4em\relax IEEE, 2015, pp. 6248--6253.

\bibitem{kim2020dynamic}
D.~Kim, S.~J. Jorgensen, J.~Lee, J.~Ahn, J.~Luo, and L.~Sentis, ``Dynamic locomotion for passive-ankle biped robots and humanoids using whole-body locomotion control,'' \emph{The International Journal of Robotics Research}, vol.~39, no.~8, pp. 936--956, 2020.

\bibitem{bellicoso2017dynamic}
C.~D. Bellicoso, F.~Jenelten, P.~Fankhauser, C.~Gehring, J.~Hwangbo, and M.~Hutter, ``Dynamic locomotion and whole-body control for quadrupedal robots,'' in \emph{2017 IEEE/RSJ International Conference on Intelligent Robots and Systems (IROS)}.\hskip 1em plus 0.5em minus 0.4em\relax IEEE, 2017, pp. 3359--3365.

\bibitem{zapolsky2014quadratic}
S.~Zapolsky and E.~Drumwright, ``Quadratic programming-based inverse dynamics control for legged robots with sticking and slipping frictional contacts,'' in \emph{2014 IEEE/RSJ International Conference on Intelligent Robots and Systems}.\hskip 1em plus 0.5em minus 0.4em\relax IEEE, 2014, pp. 3266--3271.

\bibitem{righetti2011inverse}
L.~Righetti, J.~Buchli, M.~Mistry, and S.~Schaal, ``Inverse dynamics control of floating-base robots with external constraints: A unified view,'' in \emph{2011 IEEE international conference on robotics and automation}.\hskip 1em plus 0.5em minus 0.4em\relax IEEE, 2011, pp. 1085--1090.

\bibitem{lee2022online}
J.~Lee, J.~Ahn, D.~Kim, S.~H. Bang, and L.~Sentis, ``Online gain adaptation of whole-body control for legged robots with unknown disturbances,'' \emph{Frontiers in Robotics and AI}, vol.~8, p. 788902, 2022.

\bibitem{lee2016robust}
J.~Lee, H.~Dallali, M.~Jin, D.~Caldwell, and N.~Tsagarakis, ``Robust and adaptive whole-body controller for humanoids with multiple tasks under uncertain disturbances,'' in \emph{2016 IEEE International Conference on Robotics and Automation (ICRA)}.\hskip 1em plus 0.5em minus 0.4em\relax IEEE, 2016, pp. 5683--5689.

\bibitem{farshidian2017robust}
F.~Farshidian, E.~Jelavi{\'c}, A.~W. Winkler, and J.~Buchli, ``Robust whole-body motion control of legged robots,'' in \emph{2017 IEEE/RSJ International Conference on Intelligent Robots and Systems (IROS)}.\hskip 1em plus 0.5em minus 0.4em\relax IEEE, 2017, pp. 4589--4596.

\bibitem{viceconte2022adherent}
P.~M. Viceconte, R.~Camoriano, G.~Romualdi, D.~Ferigo, S.~Dafarra, S.~Traversaro, G.~Oriolo, L.~Rosasco, and D.~Pucci, ``Adherent: Learning human-like trajectory generators for whole-body control of humanoid robots,'' \emph{IEEE Robotics and Automation Letters}, vol.~7, no.~2, pp. 2779--2786, 2022.

\bibitem{ferigo2021emergence}
D.~Ferigo, R.~Camoriano, P.~M. Viceconte, D.~Calandriello, S.~Traversaro, L.~Rosasco, and D.~Pucci, ``On the emergence of whole-body strategies from humanoid robot push-recovery learning,'' \emph{IEEE Robotics and Automation Letters}, vol.~6, no.~4, pp. 8561--8568, 2021.

\bibitem{cosner2021measurement}
R.~K. Cosner, A.~W. Singletary, A.~J. Taylor, T.~G. Molnar, K.~L. Bouman, and A.~D. Ames, ``Measurement-robust control barrier functions: Certainty in safety with uncertainty in state,'' in \emph{2021 IEEE/RSJ International Conference on Intelligent Robots and Systems (IROS)}.\hskip 1em plus 0.5em minus 0.4em\relax IEEE, 2021, pp. 6286--6291.

\bibitem{bertrand2020detecting}
S.~Bertrand, I.~Lee, B.~Mishra, D.~Calvert, J.~Pratt, and R.~Griffin, ``Detecting usable planar regions for legged robot locomotion,'' in \emph{2020 IEEE/RSJ International Conference on Intelligent Robots and Systems (IROS)}.\hskip 1em plus 0.5em minus 0.4em\relax IEEE, 2020, pp. 4736--4742.

\bibitem{mishra2021gpu}
B.~Mishra, D.~Calvert, S.~Bertrand, S.~McCrory, R.~Griffin, and H.~E. Sevil, ``Gpu-accelerated rapid planar region extraction for dynamic behaviors on legged robots,'' in \emph{2021 IEEE/RSJ International Conference on Intelligent Robots and Systems (IROS)}.\hskip 1em plus 0.5em minus 0.4em\relax IEEE, 2021, pp. 8493--8499.

\bibitem{havoutis2013onboard}
I.~Havoutis, J.~Ortiz, S.~Bazeille, V.~Barasuol, C.~Semini, and D.~G. Caldwell, ``Onboard perception-based trotting and crawling with the hydraulic quadruped robot (hyq),'' in \emph{2013 IEEE/RSJ International Conference on Intelligent Robots and Systems}.\hskip 1em plus 0.5em minus 0.4em\relax IEEE, 2013, pp. 6052--6057.

\bibitem{miki2022learning}
T.~Miki, J.~Lee, J.~Hwangbo, L.~Wellhausen, V.~Koltun, and M.~Hutter, ``Learning robust perceptive locomotion for quadrupedal robots in the wild,'' \emph{Science Robotics}, vol.~7, no.~62, p. eabk2822, 2022.

\bibitem{goodrich2007managing}
M.~A. Goodrich, T.~W. McLain, J.~D. Anderson, J.~Sun, and J.~W. Crandall, ``Managing autonomy in robot teams: observations from four experiments,'' in \emph{Proceedings of the ACM/IEEE international conference on Human-robot interaction}, 2007, pp. 25--32.

\bibitem{schermerhorn2009dynamic}
P.~Schermerhorn and M.~Scheutz, ``Dynamic robot autonomy: Investigating the effects of robot decision-making in a human-robot team task,'' in \emph{Proceedings of the 2009 international conference on multimodal interfaces}, 2009, pp. 63--70.

\bibitem{alami1998architecture}
R.~Alami, R.~Chatila, S.~Fleury, M.~Ghallab, and F.~Ingrand, ``An architecture for autonomy,'' \emph{The International Journal of Robotics Research}, vol.~17, no.~4, pp. 315--337, 1998.

\bibitem{iovino2023programming}
M.~Iovino, J.~F{\"o}rster, P.~Falco, J.~J. Chung, R.~Siegwart, and C.~Smith, ``On the programming effort required to generate behavior trees and finite state machines for robotic applications,'' in \emph{2023 IEEE International Conference on Robotics and Automation (ICRA)}.\hskip 1em plus 0.5em minus 0.4em\relax IEEE, 2023, pp. 5807--5813.

\bibitem{ghzouli2023behavior}
R.~Ghzouli, T.~Berger, E.~B. Johnsen, A.~Wasowski, and S.~Dragule, ``Behavior trees and state machines in robotics applications,'' \emph{IEEE Transactions on Software Engineering}, 2023.

\bibitem{hereid2017frost}
A.~Hereid and A.~D. Ames, ``Frost*: Fast robot optimization and simulation toolkit,'' in \emph{Proceedings of the IEEE/RSJ International Conference on Intelligent Robots and Systems}, 2017, pp. 719--726.

\bibitem{osqp}
\BIBentryALTinterwordspacing
B.~Stellato, G.~Banjac, P.~Goulart, A.~Bemporad, and S.~Boyd, ``{OSQP}: an operator splitting solver for quadratic programs,'' \emph{Mathematical Programming Computation}, vol.~12, no.~4, pp. 637--672, 2020. [Online]. Available: \url{https://doi.org/10.1007/s12532-020-00179-2}
\BIBentrySTDinterwordspacing

\end{thebibliography}

\end{document}